\documentclass[letterpaper]{article} % DO NOT CHANGE THIS
\usepackage[draft]{aaai25}  % DO NOT CHANGE THIS
\usepackage{times}  % DO NOT CHANGE THIS
\usepackage{helvet}  % DO NOT CHANGE THIS
\usepackage{courier}  % DO NOT CHANGE THIS
\usepackage[hyphens]{url}  % DO NOT CHANGE THIS
\usepackage{graphicx} % DO NOT CHANGE THIS
\usepackage{amsmath}
\usepackage{bm}
\usepackage{amsfonts}
\usepackage{natbib}  % DO NOT CHANGE THIS AND DO NOT ADD ANY OPTIONS TO IT
\usepackage{caption} % DO NOT CHANGE THIS AND DO NOT ADD ANY OPTIONS TO IT
\usepackage{algorithm}
\usepackage{algorithmic}
\usepackage{array}
\usepackage{booktabs}
\usepackage{caption}
\usepackage{cleveref}
\usepackage{array}
\usepackage{gensymb}
\usepackage{multicol}
\usepackage{amsfonts,amssymb}
\usepackage{multirow}
\usepackage{enumitem}
\usepackage{graphicx}% microtypography

\usepackage{newfloat}
\usepackage{listings}
\DeclareCaptionStyle{ruled}{labelfont=normalfont,labelsep=colon,strut=off} % DO NOT CHANGE THIS
\floatstyle{ruled}
\newfloat{listing}{tb}{lst}{}
\floatname{listing}{Listing}
\title{Training-Free Video Editing via Optical Flow-Enhanced Score Distillation}
\author{
    Lianghan Zhu\textsuperscript{\rm 1}\equalcontrib,
    Yanqi Bao\textsuperscript{\rm 1}\equalcontrib,
    Jing Huo\textsuperscript{\rm 1}\thanks{Corresponding Author.},
    Jing Wu\textsuperscript{\rm 2},
    Yu-Kun Lai\textsuperscript{\rm 2},
    Wenbin Li\textsuperscript{\rm 1},
    Yang Gao\textsuperscript{\rm 1}
}
\affiliations{
    \textsuperscript{\rm 1}Nanjing University
    \textsuperscript{\rm 2}Cardiff University\\
    \texttt{lianghanzhu@smail.nju.edu.cn, yanqibao1997@gmail.com, huojing@nju.edu.cn, WuJ11@cardiff.ac.uk, LaiY4@cardiff.ac.uk,  liwenbin@nju.edu.cn, gaoy@nju.edu.cn}

}

\usepackage{bibentry}
\begin{document}

\maketitle

%\vspace*{-1cm}
\begin{figure*}[t]
\begin{center}
\includegraphics[width=0.98\textwidth]{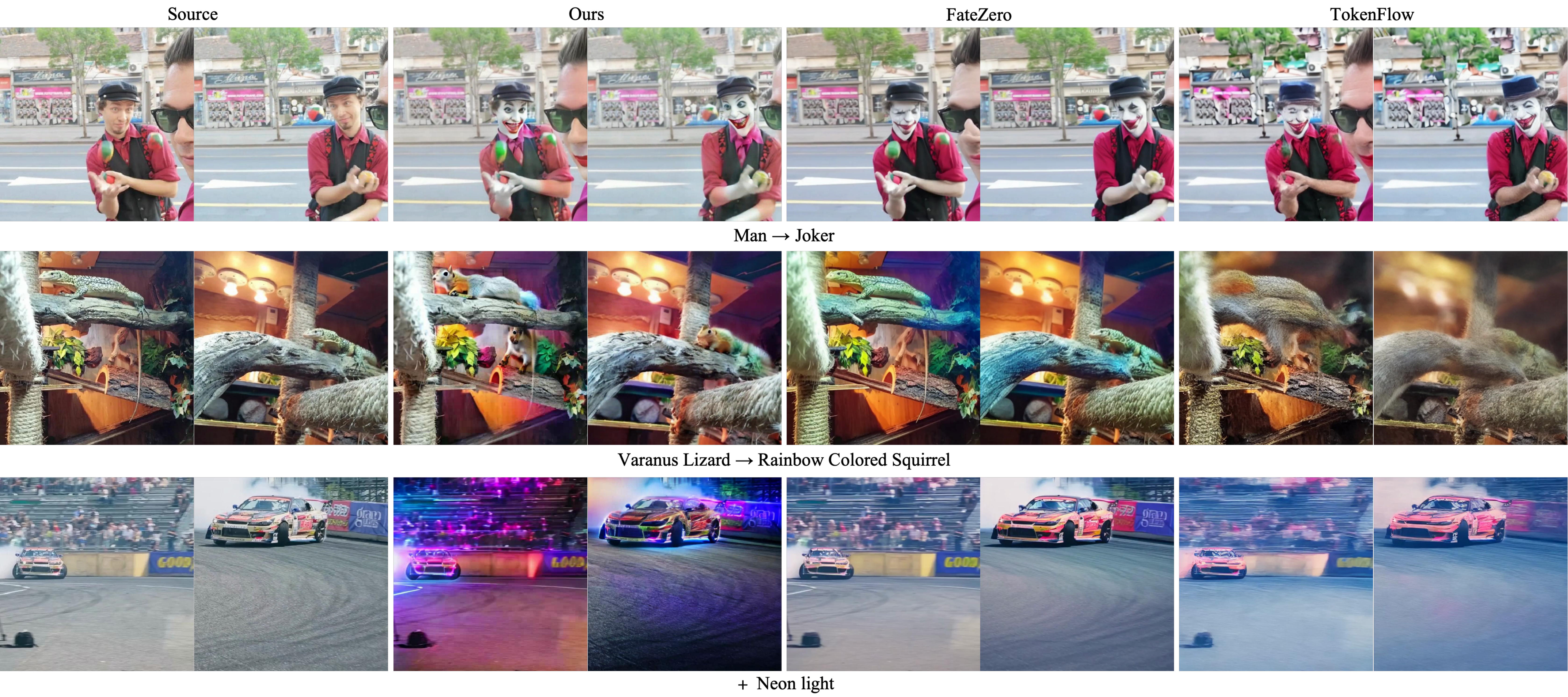}
\end{center}
\vspace{-5mm}
\caption{\textbf{Some video editing examples.} Compared to SOTA, our method achieves superior results in preserving the non-edited content of the original video, ensuring consistency and continuity in the edited results, and alignment with the target prompts.}
\label{fig1}
\vspace{-5mm}
\end{figure*}

\begin{abstract}
% The rapidly evolving field of Text-to-Video generation (T2V) has catalyzed renewed interest in train-free video editing research.
% While the application of editing prompts to guide diffusion model denoising has gained prominence, mirroring advancements in image editing, this noise-based inference process inherently compromises the original video's integrity, resulting in unintended over-editing and temporal discontinuities.
% To address these challenges, this study proposes a novel paradigm of video-based score distillation, facilitating direct manipulation of original video content.
% Specifically, distinguishing it from image-based score distillation, we propose an Adaptive Sliding Score Distillation strategy, which incorporates both global and local video guidance to reduce the impact of editing errors.
% Combined with our proposed Image-based Joint Guidance mechanism, it has the ability to mitigate the inherent instability of the T2V model and single-step sampling.
% Additionally, we design a Weighted Attention Fusion module to further preserve the key features of the original video and avoid over-editing.
% Extensive experiments demonstrate that these strategies effectively address existing challenges, achieving superior performance compared to current state-of-the-art methods.
The rapid advancement in visual generation, particularly the emergence of pre-trained text-to-image and text-to-video models, has catalyzed growing interest in training-free video editing research.
%While current approaches that mirror training-free image editing techniques—preserving original video information through video input inversion and manipulating intermediate features and attention during the inference process to achieve content editing—have demonstrated promising results, the lossy nature of the inversion process poses significant challenges in maintaining unedited regions of the video.
%%%YKL The sentence is too long and hard to follow.
Mirroring training-free image editing techniques, current approaches preserve original video information through video input inversion and manipulating intermediate features and attention during the inference process to achieve content editing. Although they have demonstrated promising results, the lossy nature of the inversion process poses significant challenges in maintaining unedited regions of the video.
Furthermore, feature and attention manipulation during inference can lead to unintended over-editing and face challenges in both local temporal continuity and global content consistency.
To address these challenges, this study proposes a score distillation paradigm based on pre-trained text-to-video models, where the original video is iteratively optimized through multiple steps guided by editing gradients provided by score distillation to ultimately obtain the target video.
The iterative optimization starting from the original video, combined with content preservation loss, ensures the maintenance of unedited regions in the original video and suppresses over-editing.
To further guarantee video content consistency and temporal continuity, we additionally introduce a global consistency auxiliary loss and optical flow prediction-based local editing gradient smoothing.
Experiments demonstrate that these strategies effectively address the aforementioned challenges, achieving comparable or superior performance across multiple dimensions including preservation of unedited regions, local temporal continuity, and global content consistency of editing results, compared to state-of-the-art methods.
\end{abstract}

\section{Introduction}
Video editing, as an interactive and controllable generation task, has consistently attracted significant attention from the AI Generated Content (AIGC) community.
These techniques enable the modification of visual content to meet specific expectations, and have found extensive applications in photography, cinematography, and artistic creation.

% The recent proliferation of Text-to-Image (T2I) and Text-to-Video (T2V) diffusion models has significantly propelled the development of this field.
The recent proliferation of Text-to-Image (T2I) and Text-to-Video (T2V) diffusion models has significantly advanced the development of downstream applications, including image editing \cite{tmm_imgedit}, image translation \cite{tmm_imgtrans}, video generation \cite{tmm_videogen, tmm_videosound}, video enhancement \cite{tmm_videoenhance}, and video editing \cite{wu2023tune,wei2023dreamvideo,customize,tcsvt_videogen,inter_video_mani,qi2023fatezero,liu2023video,geyer2023tokenflow,cong2023flatten,jeong2023ground,wang2023zero}.
Building upon these pre-trained models, current research primarily falls into two categories: zero-shot and one-shot approaches.
One-shot methods \cite{wu2023tune,wei2023dreamvideo,customize,tcsvt_videogen,inter_video_mani} fine-tune pre-trained models to learn the original video content and motion patterns, enabling video content editing through target prompt guidance during inference.
In contrast, zero-shot methods \cite{qi2023fatezero,liu2023video,geyer2023tokenflow,cong2023flatten,jeong2023ground,wang2023zero} typically leverage inversion-related \cite{hertz2022prompt, inv1, inv2, inv3, inv4} to preserve original video information, achieving video editing without additional training by manipulating intermediate features and attention computations of the U-Net during the inference process.
A common thread among these approaches is their reliance on the diffusion model's inference process starting from initial noise, progressively recovering edited video—a paradigm termed \textit{noise-based edited video generation}.
Despite leveraging inversion-related techniques to retain original video information, these methods intrinsically result in degradation of the original input structure.
Furthermore, the accumulated uncertainty during inference exacerbates this effect, %consequently 
inducing issues of interference with non-edited regions, temporal discontinuity, and content inconsistency in video editing.

An intuitive solution is to design a robust update paradigm that directly optimizes target regions in the original video (or original video latent codes) by predicting update gradients while preserving non-edited regions unchanged. 
Inspired by text-to-3D works \cite{poole2022dreamfusion}, Score Distillation Sampling (SDS) is a widely adopted method that leverages pre-trained text-to-image models to provide update guidance for initialized 3D representations, ultimately achieving 3D generation. 
Additionally, a similar concept, Delta Denoising Score (DDS) \cite{hertz2023delta}, has also been proven effective in image editing. 
These methods directly utilize update gradients provided by pre-trained models to optimize the parameters of initial representations to achieve editing. 
However, the direct application of this paradigm to video editing presents significant challenges. 
Empirical studies reveal that edited videos often exhibit over-editing and pronounced artifacts, with degradation in motion continuity and content consistency compared to the original videos. 
% We demonstrate in Fig.~\ref{fig4} the results of directly applying the DDS editing method based on the ZeroScope model to video editing.

\begin{figure}[t]
\begin{center}
\includegraphics[width=0.48\textwidth]{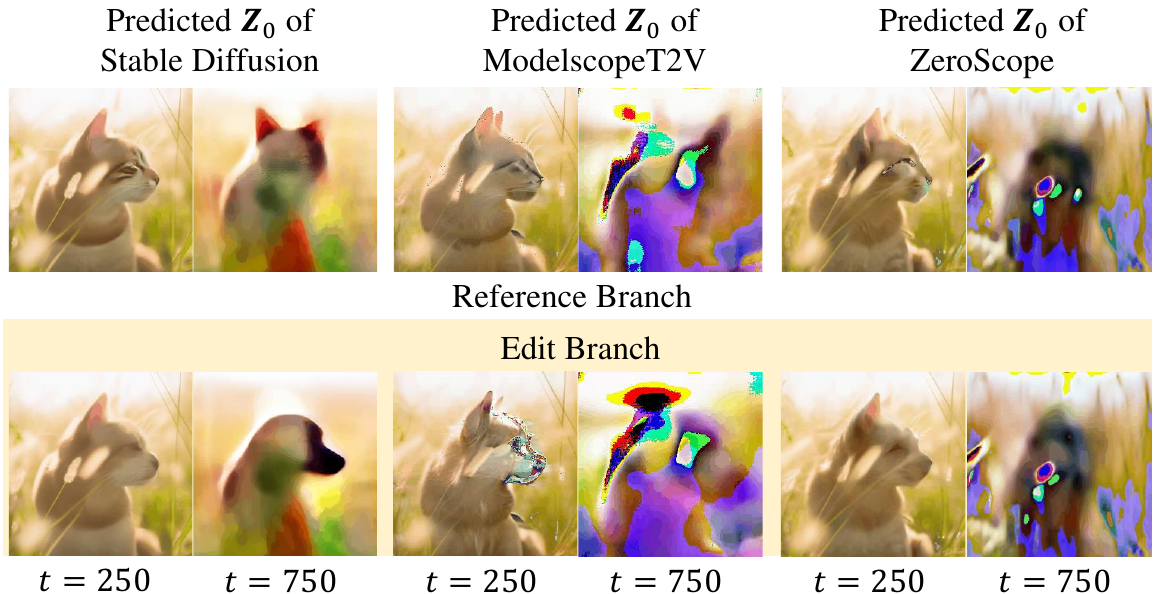}
\end{center}
\caption{
\textbf{{Visualization of One-Step Prediction of $\mathbf{Z}_0$ by Different Models at Large and Small Time Steps.}} The predicted $\mathbf{Z}_0$ is computed from the one-step predicted noise, and since visualizing $\mathbf{Z}_0$ is more meaningful, we choose to visualize $\mathbf{Z}_0$. 
}
\label{fig5}
\vspace{-6mm}
\end{figure}

We argue that the challenges stem from the weak generalization and robustness issues of T2V diffusion models, resulting in instability in the %ultimately 
computed update gradients. 
Furthermore, this paradigm computes update gradients based on single-step noise prediction from pre-trained diffusion models, which inherently exacerbates this problem.
Fig.~\ref{fig5} empirically validates this hypothesis, demonstrating that T2V diffusion models such as ModelscopeT2V and Zeroscope exhibit significant errors in single-step predicted noise used for video update gradients at larger timesteps, compared to StableDiffusion-v1.5. 
During the optimization process, biases within these update gradients progressively accumulate, ultimately leading to significant artifacts in the edited video.

To address the aforementioned challenges, this 
%This 
study introduces an optical flow-enhanced content and consistency-preserving score distillation method, designed to integrate more robust update gradients and video priors. %to address the aforementioned challenges, as illustrated in Fig.\ref{fig1}.
Specifically, we propose an improved score distillation strategy from two perspectives: editing gradient optimization and auxiliary losses.
Through a content preservation auxiliary loss between the intermediate edited video and the original video, we suppress over-editing and ensure the maintenance of non-edited regions.
By selecting anchor frames in the edited video and constraining the consistency auxiliary loss between high-level semantic features of other frames and anchor frames, we enhance the global consistency of the edited result video.
Considering the correlation between consecutive frames in videos, the editing gradients of adjacent frames should also exhibit this relationship. 
Therefore, we leverage optical flow prediction of the video to perform optical flow prediction-based smoothing operations on the editing gradients of adjacent frames, mitigating abrupt gradient fluctuations within local temporal sequences and enhancing inter-frame continuity.
Ultimately, through the comprehensive utilization of local and global information from both gradient optimization and auxiliary loss perspectives, our proposed video editing method effectively addresses the main challenges faced in the aforementioned zero-shot video editing tasks. 
% Fig.\ref{fig6} demonstrates the application of our method directly to the original video, illustrating the editing pipeline.
Comprehensive experiments conducted on text-guided real video editing tasks substantiate the efficacy of our proposed method, yielding superior results compared to alternative approaches (Fig.\ref{fig1}). In summary, our contributions are:

\begin{itemize}
% \item We propose a novel video-based optimization pipeline for zero-shot video editing through score distillation.
\item To address the issues of over-editing and global consistency in score distillation-based video editing, we introduce content preservation loss and global semantic consistency loss as auxiliary losses during the iterative optimization process.
\item We perform smoothing operations on the update gradients of adjacent frames guided by video optical flow prediction, suppressing unstable fluctuations in editing gradients and enhancing local temporal continuity.
\item Editing experiments on real videos demonstrate that our method achieves comparable or superior performance to existing methods across multiple dimensions, including alignment between the edited video and the target text prompt, continuity and consistency in the edited video, preservation of non-edited regions from the original video.
\end{itemize}

\begin{figure*}[t]
\begin{center}
\includegraphics[width=0.98\textwidth]{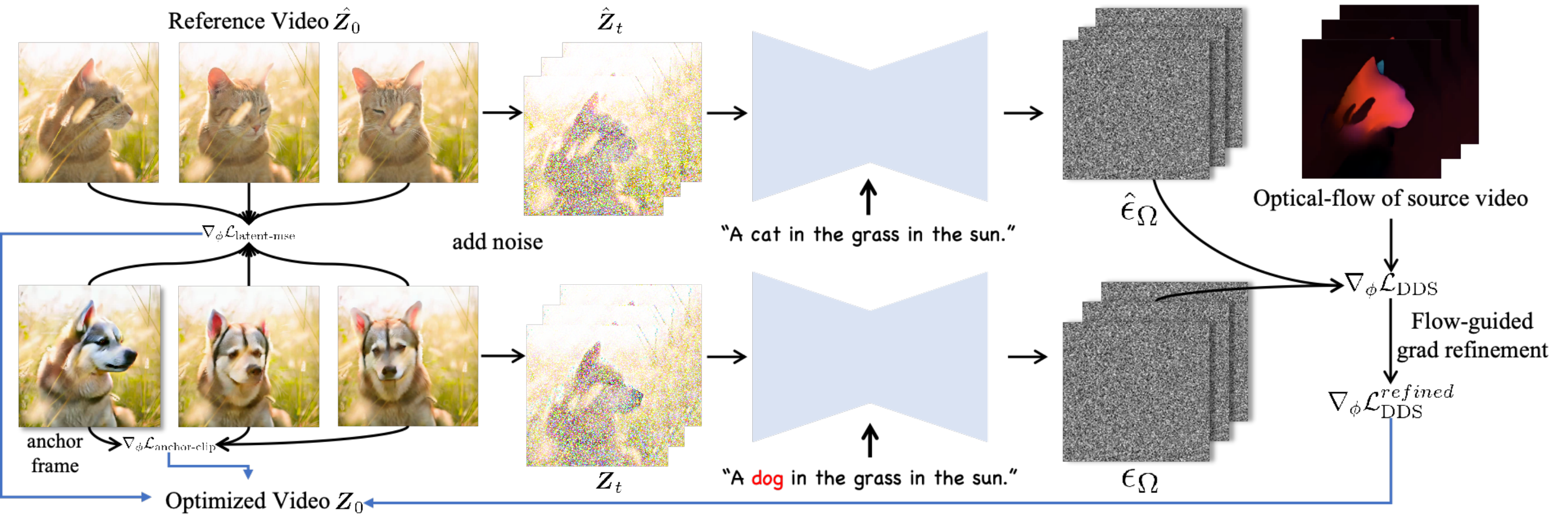}
\end{center}
\vspace{-5mm}
\caption{
\textbf{{Overview of Pipeline.}} Our pipeline comprises reference and editing branches. We employ optical flow-guided gradient refinement to enhance the continuity of gradient predictions for consecutive frame editing. Additionally, by introducing auxiliary losses for global semantic consistency and content preservation, we improve the video's global semantic consistency and maintain the original video content in non-edited regions.
}
\label{fig2}
\vspace{-5mm}
\end{figure*}

\section{Related Works}
\paragraph{Text-to-Video Generation}
Early works on conditional video generation utilized frameworks such as variational autoencoders \cite{li2018video}, generative adversarial networks \cite{yu2022generating, clark2019adversarial, saito2020train, tian2021good}, or autoregressive transformers \cite{ge2022long, hong2022cogvideo, wu2022nuwa, villegas2022phenaki}, which were trained on small domain-specific datasets. 
Despite achieving promising results, these approaches struggled to generalize to scenarios beyond the training datasets. 
Recently, numerous efforts have been made to extend the success of diffusion models in text-to-image generation to the task of text-to-video generation~\cite{Gen-2, pika}. 
By training on large-scale image and video datasets, these models have achieved remarkable generation results and controllability over the generated content. 
To accommodate the input data type of videos, the prevailing methods involve extending the convolutional and attention layers of text-to-image diffusion models along the temporal dimension, such as incorporating temporal convolutions, temporal attention layers, or introducing attention mechanisms between different frames of the video \cite{singer2022make, esser2023structure, blattmann2023align, wang2023modelscope, zhang2023show}. 
In this work, our objective is to harness the capabilities of pre-trained text-to-video diffusion models to perform zero-shot editing of original videos, given the corresponding text prompts and edited text prompts. This is achieved solely by utilizing the outputs of the pre-trained model, without subjecting it to any further training.

\paragraph{Text-based Video Editing}
With the success achieved in the field of text-based video generation\cite{hong2022cogvideo, ldm, imagen, wang2023modelscope, pika, zhang2023show}, many works have attempted to utilize T2I and T2V diffusion models for text-conditioned video editing. 
Tune-A-Video \cite{wu2023tune} pioneered one-shot video editing by fine-tuning pre-trained text-to-image diffusion models on reference videos, utilizing DDIM inversion for layout and motion guidance. Subsequent zero-shot methods emerged, employing DDIM Inversion-like techniques to preserve original video information while manipulating the inference process. Video-P2P \cite{liu2023video} adapted cross-attention control from Prompt-to-Prompt \cite{hertz2022prompt}, using a dual-branch pipeline. FateZero \cite{qi2023fatezero} manipulated intermediate features during diffusion model generation with cross-attention guidance. FLATTEN \cite{cong2023flatten} incorporated optical flow guidance. TokenFlow \cite{geyer2023tokenflow} propagated edits through feature matching, while CoDeF \cite{ouyang2023codef} decomposed videos into static content and time-dependent deformation fields for editing and propagation.

These \textit{noise-based edited video generation} methods achieve excellent video editing results, but require a complete forward diffusion and backward denoising process, often losing some original video information. 
Direct manipulation of intermediate features may also compromise the global content consistency and local temporal continuity of the resulting video.
Therefore, this study explores a paradigm that directly performs iterative optimization on the original video using editing gradients provided by pre-trained T2V diffusion models, achieving better preservation of non-edited regions in the original video, stronger global semantic consistency, and enhanced local temporal continuity through auxiliary losses and optical flow information-guided editing gradient smoothing.

\section{Preliminaries}
\paragraph{Text-conditioned Latent Video Diffusion Models.} We utilize pre-trained text-conditioned latent video diffusion models~\cite{wang2023modelscope, zeroscope} to provide guidance for video editing.
These models perform forward noise addition and reverse denoising operations in the latent space based on an autoencoder $\mathcal{E}(\cdot)$ and $\mathcal{D}(\cdot)$.
We denote a video sequence consisting of $N$ frames as $\mathbf{X}_{0}=\{\mathbf{x}_{0}^{1},..., \mathbf{x}_{0}^{N}\}$.
During the training phase, the latent input $\mathbf{Z}_{0} = \{\mathbf{z}_{0}^{1},..., \mathbf{z}_{0}^{N}\} = \mathcal{E}(\mathbf{X}_0)$ is perturbed to $\mathbf{Z}_t=\{\mathbf{z}_t^1,...,\mathbf{z}_t^N\}$ through the forward diffusion process:
\begin{equation}
\begin{aligned}\mathbf{Z}_t=\sqrt{\bar{\alpha_t}}\mathbf{Z}_0+\sqrt{1-\bar{\alpha_t}}\mathbf{\epsilon}, \mathbf{\epsilon}\sim\mathcal{N}(0,\mathbf{I})\end{aligned}, \label{eq1}
\end{equation}
for $t \in [1,T]$, where $\bar{\alpha}_t=\prod_{s=1}^t1-\beta_s$, and $\beta_s$ is the variance schedule for timestep $s$.
Subsequently, the denoising U-Net is trained to predict the added noise given condition text embedding $y$ using the following objective function:
\begin{equation}
\mathcal{L}=\mathbb{E}_{\mathcal{E}(\mathbf{X}_0),y,\mathbf{\epsilon}\sim\mathcal{N}(0,\mathbf{I}),t}\left[\|\mathbf{\epsilon}-\mathbf{\epsilon}_\theta(\mathbf{Z}_t,t,y)\|_2^2\right], \label{eq2}
\end{equation}
where $\mathbf{\epsilon}_{\theta}$ represents the 3D denoising U-Net.

\paragraph{Score Distillation for Image Editing.}
The SDS loss was first introduced in the text-to-3D work DreamFusion \cite{poole2022dreamfusion}, utilizing a pre-trained text-to-image diffusion model to provide update guidance for implicit 3D representations such as NeRF (Neural Radiance Fields) \cite{mildenhall2021nerf}.
Recent work \cite{hertz2023delta} has demonstrated the efficacy of score distillation loss in image editing tasks. 
However, due to the noise contained in the diffusion model predictions used for computing the SDS loss and the model's inherent bias, errors and noise in the obtained image editing gradient information lead to over-saturated and blurred results, causing the edited targets in the image to exhibit a cartoon-like appearance while the background becomes completely blurred.
To address this issue, DDS introduces an additional reference branch aligned with the source text prompt, in addition to the editing branch using the target text prompt. 
By utilizing the difference between the model predictions of the two branches to eliminate the noise contained in the predictions and the bias introduced by the model itself, DDS provides cleaner editing gradients.
Denote the pre-trained T2I diffusion model as $\mathbf{\epsilon}_{\theta}$.
Given the parameterized latent code $\mathbf{z}_0(\phi)$ of an image to be edited, the source text embedding $y$, and the target text embedding $y^*$.
By calculating the delta score between the model outputs of the two branches, it provides more accurate guidance for updates in the edited parts of $\mathbf{z}_0(\phi)$, while preserving the unedited parts.
Let the reference latent code be denoted as $\mathbf{\hat{z}}_0$, which is initialized as the original latent code of the image. The DDS loss can be formulated as:
\begin{equation}
\mathcal{L}_{\mathrm{DDS}}(\phi;y^*)=\left\|\mathbf{\epsilon}_\theta^w(\mathbf{z}_t(\phi),t,y^*)-\mathbf{\epsilon}_\theta^w(\mathbf{\hat{z}}_t,t,y)\right\|_2^2. \label{eq3}
\end{equation}
for $t \in [1,T]$, and $w$ represents the classifier-free guidance weight \cite{ho2022classifier} utilized in the sampling process of the diffusion model.

% \begin{figure*}[t]
% \begin{center}
% \includegraphics[width=0.98\textwidth]{optimize.pdf}
% \end{center}
% % \vspace{-3mm}
% \caption{
% \textbf{{The process of editing videos with our method.}} The images depict the gradual transformation of a jeep car into a bus/Lamborghini/truck.
% }
% \label{fig6}
% \vspace{-5mm}
% \end{figure*}

\section{Method}
We introduce our overall task pipeline in Sec.~\ref{ov}, present our proposed edit gradient correction strategy based on optical flow prediction in Sec.~\ref{grad}, and introduce our proposed content-preserving auxiliary loss and global semantic consistency auxiliary loss in Sec.~\ref{loss}.

\subsection{Overview}
\label{ov}
Given an original video $\mathbf{X}_{0}$ with its corresponding text prompt $y$ and a target text prompt $y^*$, our method directly edits the latent code $\mathbf{Z}_{0}$ of the original video to align with the target text prompt.
% As illustrated in Fig.~\ref{fig2}, we integrate the DDS loss to the video editing task and further stabilize the update gradient information used for video editing through the proposed auxiliary losses and editing gradient refinement, mitigating over-editing and enhancing the global semantic consistency and local temporal continuity of the resulting videos.
As illustrated in Fig.~\ref{fig2}, we integrate the DDS loss into the video editing framework. 
This is further augmented via a proposed content-preserving loss to maintain non-edited regions, a global semantic consistency loss to enhance overall coherence, and an optical flow-guided editing gradient refinement strategy to improve local continuity and consistency in the edited video.

\subsection{Optical Flow-Guided DDS Editing Gradient Refinement}
\label{grad}
To introduce SDS-related update paradigms in video-based editing, a natural approach is to extend DDS to temporal inputs.
Based on the advanced open-source T2V diffusion models, denoted as $\mathbf{\epsilon}_{\theta}$, video DDS loss can be expressed as:
\begin{equation}
\mathcal{L}_{\mathrm{Video-DDS}}(\phi;y^*)=\left\|\mathbf{\epsilon}_{\theta}^{w}\left(\mathbf{Z}_{t}(\phi),t,y^*\right)-\mathbf{\epsilon}_{\theta}^{w}(\mathbf{\hat{Z}}_{t},t,y)\right\|_{2}^{2}, \label{eq4}
\end{equation}
and the gradient used to update $\mathbf{Z}_0(\phi)$ can be calculated as:
\begin{equation}
\nabla_\phi\mathcal{L}_{\mathrm{Video-DDS}}=\left(\mathbf{\epsilon}_\theta^\omega\left(\mathbf{Z}_t,y^*,t \right) -\mathbf{\epsilon}_\theta^\omega\left(\mathbf{\hat{Z}}_t,y,t\right)\right)\frac{\partial\mathbf{{Z}}}{\partial\phi}. \label{eq5}
\end{equation}

% However, although pre-trained text-to-video models contain prior knowledge of global content consistency and local temporal continuity in video data, due to inherent deficiencies in the model's capabilities, the resulting edited videos still exhibit issues with local motion continuity and edited subject consistency, particularly when the edited target is temporarily occluded or moves outside the camera's field of view, where these problems may be more severe. 
However, although pre-trained text-to-video models contain prior knowledge of global content consistency and local temporal continuity, the resulting edited videos still exhibit issues with local motion continuity and edited subject consistency. 
This is particularly prominent when the edited target is temporarily occluded or moves outside the camera's field of view.
To enhance local continuity and edited subject consistency in the resulting videos, based on the correlation between adjacent frames in video data, we posit that the update gradients used for editing adjacent frames should also exhibit this correlation. 
Therefore, we refine the update gradients of adjacent frames in each iteration.

Specifically, given the original video input $\mathbf{X}_0$ containing $N$ frames, we first estimate dense inter-frame optical flow fields using the pre-trained RAFT model. 
For a frame pair $(i, j)$, the forward flow $\mathbf{f}_{i \to j} \in \mathbb{R}^{2 \times H \times W}$ and backward flow $\mathbf{f}_{j \to i} \in \mathbb{R}^{2 \times H \times W}$ are defined as:
\begin{equation}
\mathbf{f}_{i \to j} = \text{RAFT}(\mathbf{x}_i, \mathbf{x}_j), \quad \mathbf{f}_{j \to i} = \text{RAFT}(\mathbf{x}_j, \mathbf{x}_i), \label{eq6}
\end{equation}
where $\mathbf{x}_i$ and $\mathbf{x}_j$ denote the $i$-th and $j$-th frames in the input video, respectively.

In this way, we obtain forward and backward optical flows between consecutive frames in the video. 
To handle videos with more drastic changes, we additionally compute optical flow between frames with larger temporal spans beyond adjacent frames.
Based on the principle that accurate optical flow estimation should satisfy cycle consistency—i.e., mapping from frame $i$ to frame $j$ via forward flow $\mathbf{f}_{i \to j}$ and then returning via backward flow $\mathbf{f}_{j \to i}$ should arrive at the original position—we apply cycle consistency filtering to the bidirectional optical flow to obtain more reliable flow estimates. 
Only the flow information that passes this filtering is used to guide the subsequent gradient refinement operations.
Finally, we downsample the filtered optical flow fields to the latent space for subsequent operations.

Given the DDS gradient $\nabla_{\mathbf{z}_i} \mathcal{L}_{\text{DDS}}$ for updating the latent features corresponding to frame $i$, we propagate gradient information from neighboring frames to the current frame through the optical flow field. 
Specifically, for frame $j = i \pm h$, where $h$ is a predefined hop count that controls the temporal range of optical flow influence, we define the gradient warping operation as:
\begin{equation}
\widetilde{\nabla}_{\mathbf{z}_{j \to i}} = \mathcal{W}\left(\nabla_{\mathbf{z}_j} \mathcal{L}_{\text{DDS}}, \mathbf{f}_{j \to i}\right).
\label{eq7}
\end{equation}
Finally, we perform weighted fusion of the original DDS gradient for each frame with the warped gradients from other frames to obtain the refined gradient. For frame $i$, the refined gradient $\nabla_{\mathbf{z}_i}^{\text{refined}}$ is computed as:
\begin{equation}
\nabla_{\mathbf{z}_i}^{\text{refined}} = (1 - \alpha) \cdot \nabla_{\mathbf{z}_i} \mathcal{L}_{\text{DDS}} + \alpha \cdot \sum_{h=1}^{H_{\max}} \omega_h \cdot (\widetilde{\nabla}_{\mathbf{z}_{i+h \to i}} + \widetilde{\nabla}_{\mathbf{z}_{i-h \to i}}),
\label{eq8}
\end{equation}
where $\alpha$ controls the strength of gradient refinement, and $\omega_h$ controls the fusion weight for frames at different temporal distances from frame $i$.

Our proposed optical flow-guided gradient refinement strategy effectively addresses the temporal continuity issue of DDS in video editing by explicitly modeling inter-frame motion relationships. 
As demonstrated in the ablation study (Fig.~\ref{fig4}), this mechanism enhances the continuity and consistency between consecutive frames, especially in cases where the edited object is occluded.

\subsection{Content Preservation and Global Semantic Consistency Auxiliary Losses}
\label{loss}
Although the editing gradients computed based on model predictions are primarily applied to regions requiring editing under the guidance of the target text, they can still introduce unwanted or unstable over-editing in both edited and non-edited regions due to the uncertainty in model predictions and the noise they contain. 
Such errors accumulate over multiple iterations and ultimately have a significant impact on the results.

Considering that the update gradients causing over-editing are relatively smaller compared to the effective update gradients guiding the editing, we introduce a content preservation loss at each iteration to suppress the influence of noisy update gradients on the results. 
At each iteration, we introduce the MSE loss between the latent $\mathbf{z}$ in the editing branch and the latent $\hat{\mathbf{z}}$ in the reference branch as an auxiliary content preservation loss:
\begin{equation}
\mathcal{L}_{\text{preserve}} = \text{MSE}(\mathbf{z}, \hat{\mathbf{z}}) = \|\mathbf{z} - \hat{\mathbf{z}}\|_2^2.
\label{eq9}
\end{equation}
The editing gradients, which are derived from the single-step noise prediction of a pre-trained text-to-video diffusion model, may introduce undesirable updates in regions meant to be preserved. 
To mitigate this, we incorporate a per-step content preservation loss with respect to the original video frame. 
Given that the editing gradients in non-target regions are substantially weaker, this loss dominates there to maintain original content, while the DDS-based editing gradients continue to drive changes in the target regions.
As demonstrated in Fig.~\ref{fig4}, the ablation analysis confirms that this auxiliary loss effectively reduces the impact of noise and over-editing in the editing gradients on the results.

Although the pre-trained text-to-video foundation model we employ contains prior knowledge of global semantic consistency, we still observe semantic inconsistencies in edited objects between the beginning and end portions of videos in some cases. 
To enhance the global semantic consistency of edited videos, we introduce an anchor frame-based global semantic consistency loss. 
We select one frame of the video as the anchor frame (more frames can also be selected based on video content if the content in the video undergoes significant changes over time). 
At each iteration, we decode the latent $\mathbf{z}$ of the video being edited to pixel space, then use a pre-trained CLIP image encoder to extract CLIP visual features of the anchor frame $\mathbf{x}_{\text{anchor}}$ and other frames $\mathbf{x}_i$ ($i \neq \text{anchor}$), and compute the cosine similarity loss between the CLIP visual features of other frames and the anchor frame:
\begin{equation}
\mathcal{L}_{\text{semantic}} = \frac{1}{N-1} \sum_{i \neq \text{anchor}} \left(1 - \frac{\text{CLIP}(\mathbf{x}_i) \cdot \text{CLIP}(\mathbf{x}_{\text{anchor}})}{\|\text{CLIP}(\mathbf{x}_i)\| \|\text{CLIP}(\mathbf{x}_{\text{anchor}})\|}\right).
\label{eq10}
\end{equation}

Finally, the total loss at each iteration consists of the video DDS loss, content preservation auxiliary loss, and global semantic consistency auxiliary loss:
\begin{equation}
\mathcal{L}_{\text{total}} = \mathcal{L}_{\text{DDS}} + w_1 \mathcal{L}_{\text{preserve}} + w_2 \mathcal{L}_{\text{semantic}}
\label{eq11}
\end{equation}
where $w_1$ and $w_2$ are weights that control the content preservation loss and global semantic consistency loss, respectively.
These auxiliary losses effectively suppress over-editing and enhance the global semantic consistency of the resulting videos.

\begin{figure*}[!htbp]
\begin{center}
\includegraphics[width=0.98\textwidth]{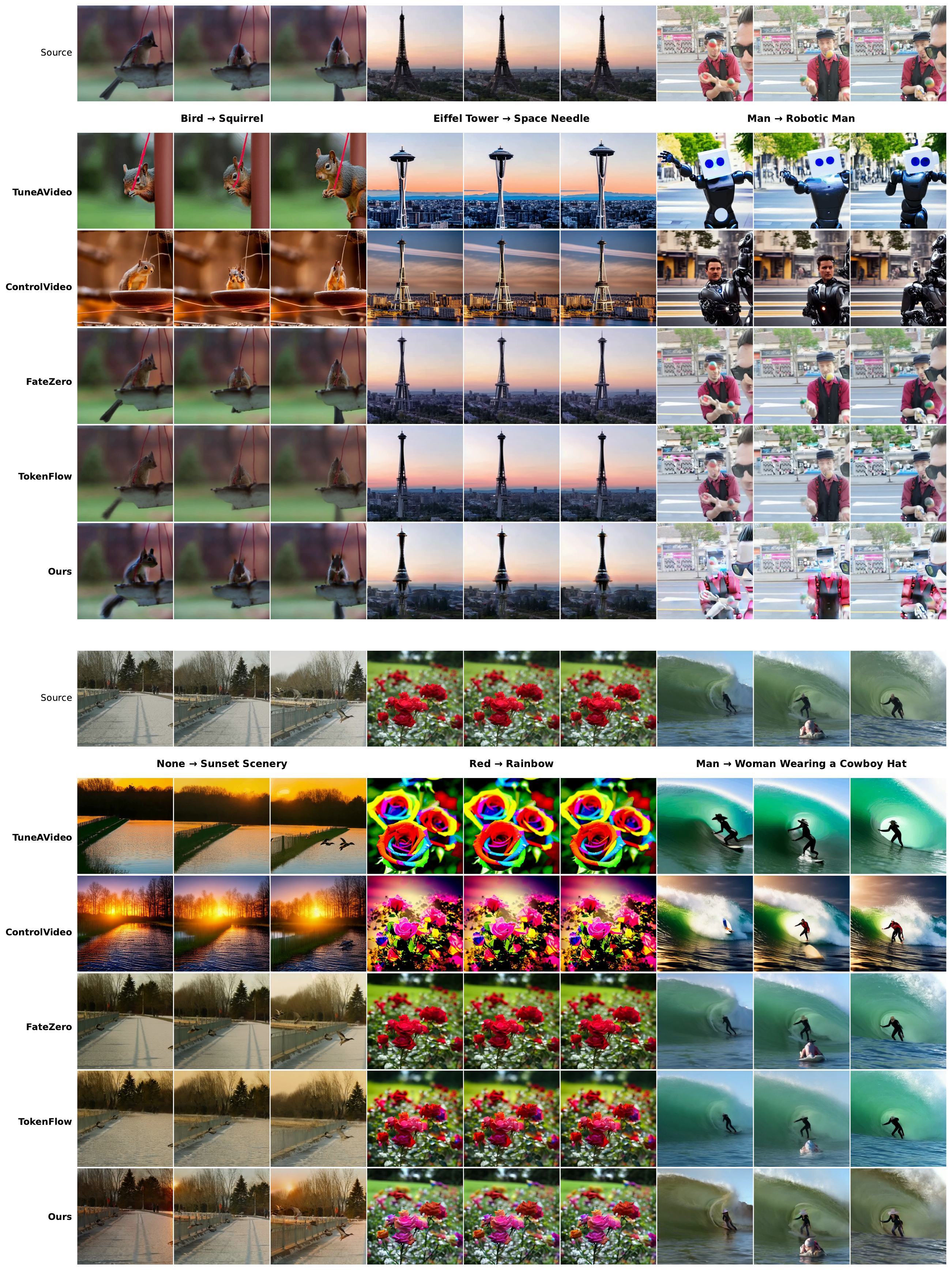}
\end{center}
\vspace{-5mm}
\caption{
\textbf{{Qualitative Comparison.}} Our method achieves a good balance between editing and preserving unedited information in the original video. Additionally, it possesses a certain capability for geometric shape editing. 
}
\label{fig3}
\vspace{-5mm}
\end{figure*}

\section{Experiments}
We introduce our experimental setup in Sec.~\ref{setting} and present visual comparisons between our method and baseline methods in Sec.~\ref{quali}. In Sec.~\ref{quant}, we describe the definitions of the quantitative metrics used, discuss the limitations of existing quantitative metrics, and report the quantitative comparison results between our method and baseline methods. In Sec.~\ref{ablat}, we validate the effectiveness of each proposed improvement. Finally, in Sec.~\ref{basemodel}, we verify the compatibility of our method with different text-to-video foundation models.

\subsection{Experiments Setting}
\label{setting}
\paragraph{Data Preparation}
We constructed over 100 text-guided editing cases for real videos, with each case consisting of three components: the original video, the text prompt corresponding to the original video, and the target editing text prompt. 
The videos were obtained from open-source datasets including DAVIS \cite{davis}, WebVid \cite{webvid}, and YouTube \cite{abu2016youtube}, from which 32 consecutive frames were selected and extracted for construction.
The text prompt corresponding to the original video is a one-sentence description of the video content, while the target editing text prompt is constructed by modifying certain phrases in the original text prompt, or adding additional descriptions.
These cases involve editing the objects, backgrounds, and styles in videos.

\paragraph{Experiment Details}
We employed the open-source pre-trained text-to-video model ZeroScope as our base model.
For optimizing the latent code of one video, we employed the SGD optimizer with a learning rate of 0.1 for 300 steps, taking approximately 6 to 7 minutes on a single A6000 GPU.
For video optical flow prediction, we utilized the large version of the RAFT model provided by the torchvision library, with 20 iterative updates to achieve a balance between efficiency and accuracy.
In the global semantic consistency loss, we employed \textit{openai/clip-vit-base-patch32} as the high-level semantic visual feature extractor.

\paragraph{Baselines}
To evaluate the performance of our proposed method, we evaluated our method against five SOTA video editing methods:
Tune-A-Video \cite{wu2023tune} fine-tunes the denoising U-Net on a single video to learn the structural integrity of the original video.
FateZero \cite{qi2023fatezero} achieves zero-shot video editing via DDIM Inversion and attention manipulation.
% FLATTEN \cite{cong2023flatten} uses an external optical flow detection model for accurate attention manipulation.
TokenFlow \cite{geyer2023tokenflow} ensures consistency by modifying keyframes and propagating changes throughout the video.
ControlVideo \cite{zhang2023controlvideo} leverages depth and edge information from the original video to control editing results.

\begin{figure*}[t]
\begin{center}
\includegraphics[width=0.98\textwidth]{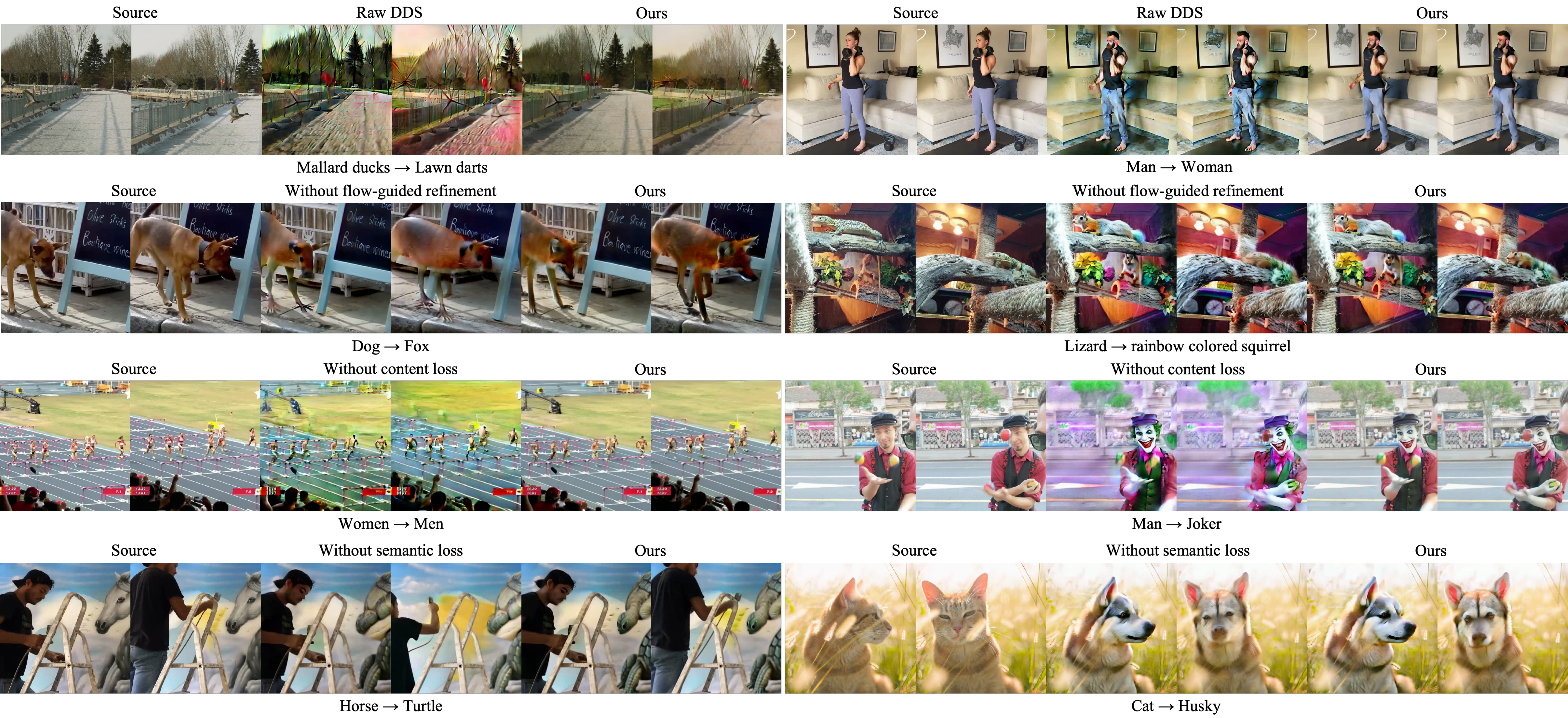}
\end{center}
\caption{
\textbf{{Abalation of optical flow-guided gradient correction strategy, content preservation auxiliary loss, and global semantic consistency auxiliary loss.}} In most cases, using all the components we propose achieves the best balance between editing and preservation, and achieves the highest edited video quality. 
}
\label{fig4}
\vspace{-6mm}
\end{figure*}

\subsection{Qualitative Experiment}
\label{quali}
We conduct qualitative comparisons between our method and baseline methods on constructed real-world video editing cases.
Fig.~\ref{fig3} presents the editing results of our method and baseline methods on several representative examples.
Tune-A-Video achieves partial preservation of original video information through DDIM Inversion and generates videos that reasonably align with the target text prompts. 
However, due to the lack of temporal priors in the base model and strong constraints related to the original video, the edited results exhibit poor temporal consistency and severe flickering artifacts. 
Moreover, the edited videos lose substantial information from the source video, with minimal preservation of non-edited regions.
ControlVideo leverages conditions such as depth information from the original video to achieve layout control in the edited results, maintaining similar layout and motion patterns to the source video. 
However, this preservation is limited to high-level semantics and still lacks consistency with the original video at the detail level. 
By utilizing only depth and other semantically-weak conditions from the source video for control, the editing becomes more text-aligned while reducing the influence from the editing targets in the original video.
FateZero not only obtains the initial noise corresponding to the original video through DDIM Inversion but also stores intermediate features from each step of the inversion process. 
By utilizing these intermediate features, it achieves better preservation of original video information. 
However, this comes at the cost of limited editing capability, typically restricted to modifying object surface textures, with difficulties in editing geometric shapes or colors.
TokenFlow faces similar limitations, struggling to support modifications of geometric shapes and significant color variations.
In summary, existing methods face trade-offs among preventing over-editing, preserving non-edited content of the original video, and achieving edits that align with the target text.
Methods with stronger capability in preserving original video information typically exhibit weaker editing capability, struggling to modify attributes beyond color and texture of the edited objects, such as FateZero and TokenFlow.
The preservation of original video content and layout also enables edited results to maintain stronger temporal continuity.
While methods with fewer constraints on original video information generate editing results that better align with the target text but typically fail to preserve non-edited regions, such as ControlVideo and Tune-A-Video.
Our method achieves the best comprehensive results in preserving non-edited regions and partial information of edited objects in the video, as well as maintaining local temporal continuity and global semantic consistency.
% Additional qualitative comparison results are provided in the Appendix.

\begin{table}[!t]
\caption{Auto Metrics Comparison of Different Methods\label{tab_qua}}
\centering
\begin{tabular}{|c||c|c|c|c|}
\hline
method & Pres & CLIP$_{img}$ & CLIP$_{text}$ & VQAScore\\
\hline
Tune-a-Video & 0.1971 & 0.9432 & 0.3242 & 0.4439\\
\hline
ControlVideo & 0.0587 & 0.9773 & 0.3103 & 0.3386\\
\hline
Fatezero & 0.6628 & 0.9776 & 0.2929 & 0.3235\\
\hline
Tokenflow & 0.5704 & 0.9791 & 0.3165 & 0.3081\\
\hline
Ours & 0.7215 & 0.9660 & 0.3111 & 0.3877\\
\hline
\end{tabular}
\end{table}

\begin{table}[!t]
\caption{User Study Comparison of Different Methods\label{tab_user}}
\centering
\begin{tabular}{|c||c|c|c|c|c|}
\hline
method & Align & Cons \& Cont & Pres & Avg\\
\hline
Tune-a-Video & 0.5377 & 0.0837 & 0.0882 & 0.2365\\
\hline
ControlVideo & 0.5618 & 0.1883 & 0.2261 & 0.3254\\
\hline
Fatezero & 0.4417 & 0.4792 & 0.4958 & 0.4722\\
\hline
Tokenflow & 0.5129 & 0.4416 & 0.4589 & 0.4711\\
\hline
Ours & 0.7242 & 0.6182 & 0.6494 & 0.6639\\
\hline
\end{tabular}
\end{table}

\subsection{Quantitative Experiment}
\label{quant}
\paragraph{Auto Metrics}
Following established practices in the literature, we employ the CLIP model \cite{radford2021learning} for automatic evaluation, utilizing two primary metrics: CLIP Text Alignment Score and CLIP Frame Consistency Score.
It should be noted, however, that existing automatic metrics possess inherent limitations.
The CLIP Text Alignment Score, defined as the mean cosine similarity between CLIP image embeddings of edited video frames and the CLIP text embedding of the target prompt, quantifies text-video alignment.
Nevertheless, a single textual description cannot encapsulate the informational richness and nuanced details of the original video, rendering this metric insufficient for assessing preservation of non-edited content.
The CLIP Frame Consistency Score, defined as the mean cosine similarity between CLIP image embeddings across video frames, evaluates semantic consistency.
However, as CLIP features are high-level and predominantly semantic, this metric fails to capture local temporal continuity and consistency in low-level visual attributes.
We also introduce VQAScore \cite{vqa}, a metric specifically proposed for text-to-video tasks that better aligns with human subjective visual perception, as an evaluation metric for video editing performance. 
This metric takes the edited video and the target text prompt as inputs, and can simultaneously evaluate both the overall quality of the generated video that better aligns with human subjective assessment and the alignment with the target text. 
Concerning preservation of non-edited regions, frame-wise low-level feature similarity computation between edited and original videos is problematic in the absence of precise per-frame editing masks. 
While high similarity suggests better preservation, it inherently conflicts with achieving effective text-aligned edits.
Therefore, in evaluating the preservation capability of original video information, we avoid directly measuring the low-level feature similarity between the edited and original videos. 
Instead, we assess whether the video editing preserves the color characteristics of the original video by computing and comparing the color histogram similarity between corresponding frames of the source and edited videos.
Furthermore, no widely accepted automatic evaluation method currently exists for assessing local temporal continuity in such tasks.
Quantitative results for these automatic metrics are presented in Table.~\ref{tab_qua}.
Due to the trade-off between preserving original video information and achieving edits that better align with the target text, as discussed in the previous qualitative experiments, no method consistently achieves optimal performance across all quantitative metrics. 
Overall, however, our method strikes a better balance between preserving original video information and generating edits that conform to the target text.

\begin{figure*}[t]
\begin{center}
\includegraphics[width=0.98\textwidth]{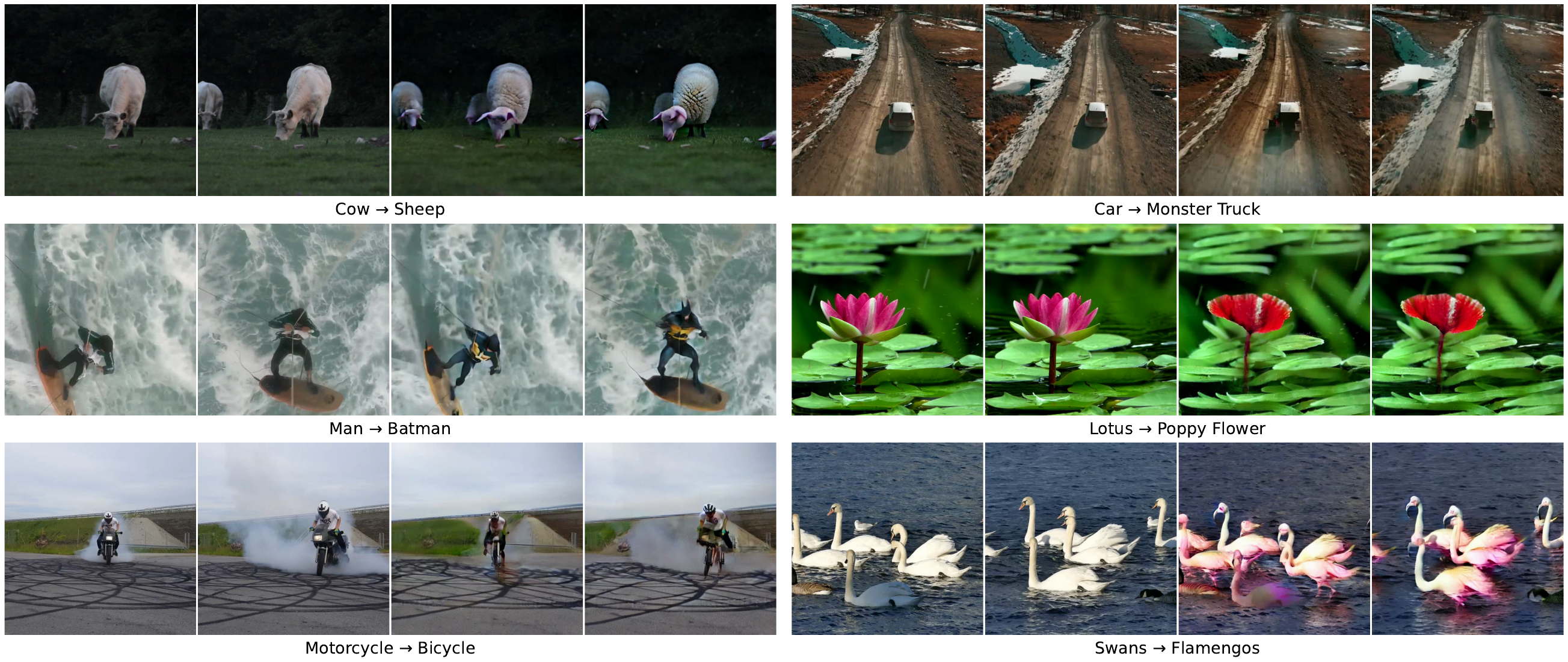}
\end{center}
% \vspace{-3mm}
\caption{
\textbf{{Results using ModelscopeT2V as the text-to-video foundation model.}} Our method is compatible with different text-to-video foundation models.
}
\label{fig7}
\vspace{-5mm}
\end{figure*}

\paragraph{User Study}
Given the limitations of automatic evaluation metrics, we conducted a user study to assess dimensions that are challenging to quantify automatically yet critical for video editing quality, thereby obtaining evaluations consistent with human visual perception.
40 volunteers were recruited to participate in the user study.
Volunteers were instructed to evaluate edited videos along three dimensions: (1) alignment between the edited video and the target text prompt; (2) continuity and consistency in the edited video; (3) preservation of non-edited regions from the original video.
For each editing case, volunteers are required to rate three dimensions on a scale from 1 to 5. To eliminate the differences caused by varying rating scales among different volunteers, we first normalize all rating results from each volunteer to the range of [0, 1]. 
We then calculate the average score across all examples for each method in the three dimensions as the final evaluation results to report.
As shown in Table.~\ref{tab_user}, our method achieves the highest scores in human subjective evaluation across all three evaluation dimensions. 
Furthermore, the scores across different dimensions also reflect the trade-off between preserving original video information and achieving edits that better align with the text prompts, as mentioned earlier for different methods.
Since the non-edited portions in the text prompts paired with videos are not perfectly aligned with the corresponding content in the videos, our method's stronger capability in preserving the non-edited content of the original video actually suppresses the alignment of these non-edited portions toward their corresponding text descriptions. 
This results in our method not achieving the highest scores on the CLIP$_{text}$ and VQAScore metrics. 
However, human volunteers can understand the intended editing from the source and target text prompt pairs, thereby excluding the influence of irrelevant descriptions in the text prompts and providing a more authentic assessment of the alignment between the edited video and the target text prompt. 
Consequently, our method achieves the highest score in the Alignment evaluation dimension.
Although computing the CLIP$_img$ scores between adjacent frames is intended to measure the consistency and continuity of video frames, the output features extracted by the CLIP Image encoder are overly abstract and low-dimensional, primarily capturing high-level semantic consistency while struggling to capture low-level features such as fine-grained continuity and consistency.
Tune-A-Video and ControlVideo, which exhibit severe flickering artifacts in their editing results, also achieve high scores on this metric, exposing its limitations.
This also explains why our method, despite achieving ordinary performance on the CLIP$_img$ metric, demonstrates the best performance in human evaluation of inter-frame consistency and continuity.

\subsection{Ablation Study}
\label{ablat}
We conduct ablation studies to validate the effectiveness of our proposed optical flow-guided gradient correction strategy, content preservation auxiliary loss, and global semantic consistency auxiliary loss.

\textbf{Optical Flow-Guided Gradient Correction Strategy} Leveraging prior knowledge about temporal correlations between consecutive frames, we employ optical flow predictions to guide the correction of DDS editing gradients, thereby improving temporal continuity in edited videos.
The experimental results in Fig.~\ref{fig4} demonstrate that the gradient correction guided by multi-frame optical flow enhances the consistency of motion and subjects across consecutive frames, particularly when the edited target is occluded or moves outside the camera's field of view. 
This strategy significantly improves subject consistency in such scenarios. 
As shown in the left example of the second row, when the edited target is occluded by a sign and its head is blocked due to turning away during the video sequence, the editing results without the optical flow-based strategy exhibit significantly lower quality compared to those employing this strategy. 
In the right example of the second row, although consistent editing is achieved in the first frame regardless of whether this strategy is used, the subsequent motion causes the squirrel to be occluded by the wooden stick on the left side of the first frame. 
Consequently, without this strategy, the squirrel loses the appearance established in the first frame in subsequent frames.
Our gradient correction strategy effectively enhances temporal coherence in the edited results.

\textbf{Content Preservation Auxiliary Loss} To mitigate over-editing and preserve information in non-edited regions, we introduce a content preservation loss that reduces the impact of noise and errors in video DDS editing gradients. Fig.~\ref{fig4} illustrates that The Content Preservation Auxiliary Loss effectively mitigates the impact of errors in the editing gradients on the editing results. 
In the results shown in the third row, the outcomes without Content Preservation Auxiliary Loss exhibit numerous color artifacts.
The content preservation auxiliary loss effectively prevents over-editing and substantially improves output video quality.

\textbf{Global Semantic Consistency Auxiliary Loss} While pretrained text-to-video models incorporate architectures designed for temporal modeling, edited objects may still exhibit semantic inconsistencies due to limited model capacity. To strengthen global semantic consistency constraints, we introduce an auxiliary loss based on the CLIP visual encoder. As evident from Fig.~\ref{fig4}, the results in the fourth row demonstrate that, in certain details, the Global Semantic Consistency Auxiliary Loss is more conducive to maintaining high-level semantic consistency across the entire video at a global scale.
This auxiliary loss enhances global consistency of edited subjects across the video sequence.

\subsection{Compatibility With Other Foundation Models}
\label{basemodel}
Although the visual and quantitative results reported in our paper are based on the ZeroScope foundation model, it should be noted that our method is readily compatible with other diffusion-based text-to-video models.
We present several results using ModelscopeT2V as the text-to-video diffusion model in Fig.~\ref{fig7}, demonstrating the compatibility of our method with different foundation models.

\begin{figure}[t]
\begin{center}
\includegraphics[width=0.48\textwidth]{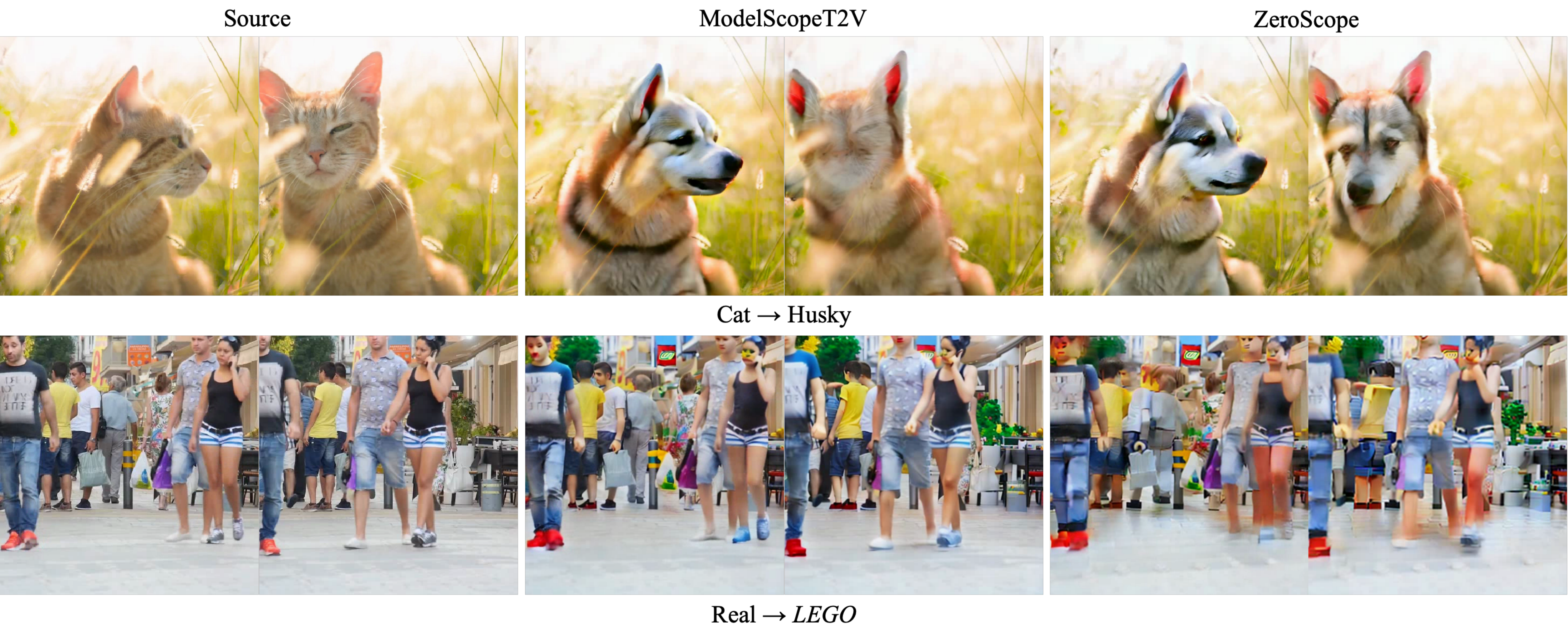}
\end{center}
\caption{
\textbf{{Comparative Visualization Across Foundation Models.}} The disparity in foundational model capabilities leads to inferior outcomes from ModelScopeT2V compared to ZeroScope on same cases. 
}
\label{fig8}
\vspace{-6mm}
\end{figure}

\section{Conclusion and Limitations}
We migrate the Delta Denoising Score approach to video editing tasks and optimize the editing performance by combining the proposed auxiliary losses and gradient refinement methods, providing a novel score distillation-based paradigm for video editing.
The content preservation loss suppresses over-editing and ensures the invariance of non-edited regions in the original video by constraining the consistency between the video being edited and the original video.
The global semantic consistency loss enhances the global semantic consistency in the resulting video by constraining the consistency of high-level semantic information between other frames and anchor frames in the video.
The optical flow prediction-guided editing gradient refinement smooths the editing gradients at corresponding positions in consecutive frames, promoting local temporal continuity in the resulting video.
Our experimental results on real video editing validate the efficacy of our paradigm and the proposed auxiliary losses and gradient refinement strategies in video editing tasks.

However, such methods currently face the same limitations shared by other zero-shot editing approaches, namely being constrained by the reconstruction and generation capabilities of the pre-trained foundation generative models themselves. 
Fig.~\ref{fig8} presents the experimental results obtained using ModelScopeT2V and ZeroScope as T2V foundation models on the same cases. 
Due to differences in the capabilities of the pre-trained models, in Case 1 where a cat is edited into a husky, the output from ModelScopeT2V only blurs the original cat's head in the latter part of the sequence, failing to achieve an edit consistent with the target text. 
In Case 2, which involves editing a real-world scene into LEGO bricks, the results from ModelScopeT2V also show little change aligning with the target description. 
This suggests that ModelScopeT2V appears to lack sufficient knowledge of the characteristic features of a LEGO brick scene.
The effectiveness of score distillation methods depends on the premise that pre-trained generative diffusion models can provide update gradients aligned with the target text prompt, despite inherent biases. 
Nevertheless, the limited robustness and generalization capabilities of current open-source T2V diffusion models restrict the performance ceiling of such methods.

\bibliography{aaai25}

\end{document}